\documentclass[11pt]{article}
\usepackage[T1]{fontenc}
\usepackage[utf8]{inputenc}
\usepackage{lmodern}
\usepackage[margin=1in]{geometry}
\usepackage{amsmath,amssymb,amsthm,mathtools}
\usepackage{graphicx,booktabs,tabularx,array}
\usepackage[round,authoryear]{natbib}
\usepackage{microtype}
\usepackage{xcolor}
\usepackage[colorlinks=true,linkcolor=blue!45!black,citecolor=blue!45!black,urlcolor=blue!45!black]{hyperref}
\usepackage{enumitem}
\setlist{itemsep=2pt,topsep=4pt}
\newtheorem{theorem}{Theorem}[section]
\newtheorem{proposition}[theorem]{Proposition}
\newtheorem{corollary}[theorem]{Corollary}

\theoremstyle{definition}

\newtheorem{example}[theorem]{Example}
\theoremstyle{remark}

\newcommand{\Prob}{\mathbb P}
\newcommand{\E}{\mathbb E}
\newcommand{\ind}{\mathbf 1}
\newcommand{\supp}{\operatorname{supp}}
\newcommand{\cX}{\mathcal X}
\newcommand{\cF}{\mathcal F}
\newcommand{\cG}{\mathcal G}
\newcommand{\cR}{\mathcal R}
\newcommand{\cA}{\mathcal A}
\newcommand{\cI}{\mathcal I}
\newcommand{\cC}{\mathcal C}

\newcommand{\CodingModels}{96}
\newcommand{\EncoderPartitions}{203}
\newcommand{\DecoderChecks}{19488}
\newcommand{\HazardModels}{100}
\newcommand{\HazardPolicies}{616}
\newcommand{\ChannelModels}{144}
\newcommand{\FactorModels}{10}
\newcommand{\FactorQueries}{1213}
\newcommand{\PrescriptionSequences}{4096}
\newcommand{\PrescriptionValue}{16/25}

\title{Poincar\'e Meets Bellman:\\
Revisable Memory, Operational Quotients, and\\
Evidence-Supported Learning in Changing Environments}
\author{Xin Li\\\small Department of Computer Science, University at Albany}
\date{October 1, 2026}
\begin{document}
\maketitle
\begin{abstract}
Memory consolidation determines both what a learner can do now and which
changes remain implementable later. We develop a finite-model synthesis of
operational state abstraction and optimal control under the
stability--evidence--revision (SER) framework. ``Poincar\'e meets Bellman''
names two complementary roles: qualitative dynamics identifies reusable
action--response structure, and dynamic programming prices acquisition,
retention, reuse, merging, and forgetting. Recurrence enters separately
through the timing and value of future demands. We distinguish active
quotient merging from historical information erasure, characterize exact
repair by zero-error functional coding and causal migration, and derive a
Bellman recursion over the joint law of hidden state and complete deployed
memory. A first-return model yields an explicit retention rule. Conditional
results show how factor sharing avoids enumerating combinations and how
independent informative observations improve identification, while leaving
some zero-error evidence budgets unchanged. Finite enumerations verify the
coding and retention calculations. The synthesis gives an exact benchmark
for specified finite models, without claiming universal recurrence,
bounded-memory open-ended learning, or tractable global planning.
\end{abstract}

\section{Learning behavior and preserving future learning}
\label{sec:framing}

A learner in a changing world encounters both novelty and recurrence.
Some experiences reveal mechanisms worth reusing. Others expose conditions
under which an established mechanism fails. Memory must support both kinds
of learning under finite storage and computation budgets. Retaining every
experience is expensive, but preserving current performance alone can erase
the distinctions needed for a later repair.

The \emph{stability--evidence--revision} (SER) problem makes three
requirements explicit: preserve declared competence, retain or acquire the
information needed for useful changes, and commit to a structural change
under a declared evidential criterion. We use \emph{revisable learning} for
the resulting problem of organizing experience while preserving selected
capacities for future revision. Reinforcement learning supplies an action
and return objective. Revisable learning adds decisions about the
information and structure through which future action selection and
learning will operate. A single controller can address both objectives.
Representation learning and memory already occur within RL; the proposed
distinction concerns the objective being optimized.

\subsection{The two roles in Poincar\'e meets Bellman}

Poincar\'e's qualitative study of dynamics motivates reasoning about
invariant organization and future behavior beyond individual trajectories
\citep{poincare1890}. Here the operational question is which histories or
problem instances can share an interface because their relevant future
action consequences agree. Exact conjugacy can establish such agreement
in suitable deterministic systems, but general transfer does not require
topological conjugacy. We work with \emph{operational equivalence}, defined
relative to a specified response, action, and continuation family.

Bellman's principle prices decisions by immediate cost and optimal
continuation value \citep{bellman1954}. Applied to memory, the continuation
problem depends on earlier retention choices. Erasing a witness can remove
an entire class of future repairs. Keeping a witness can have positive
value even when the witness changes no present prediction.

The phrase \emph{Poincar\'e meets Bellman} denotes that synthesis. The
qualitative-dynamics role identifies candidate reusable structure. The
control role determines which structure and evidence are worth acquiring,
retaining, revising, and retiring. The phrase is a research framing, not
the name of an established joint theorem. Poincar\'e recurrence supplies a
separate classical result under measure-preservation assumptions; neither
stationarity nor the existence of a learned hierarchy establishes those
assumptions.

\subsection{What the finite synthesis establishes}

The paper develops four connections. First, quotient factorization and
controlled state abstraction explain when merging preserves a specified
competence or decision problem. Second, coding and migration separate a
repair's information requirement from its implementability after deletion.
Third, finite prescription optimization compares complete representation
and evidence trajectories. Fourth, explicit factor and observation models
identify conditions under which additional dimensions support reuse or
identification.

Supporting results are attributed where classical. The controlled quotient
and Bellman arguments specialize standard abstraction and dynamic
programming principles. The revision budget specializes zero-error
side-information and functional coding. The paper's organizing contribution
is the joint treatment of operational organization, historical access,
recurrence value, and supported commitment. The first-return calculation
and finite examples make those connections concrete. No biological
mechanism or universal architecture optimum is inferred from the finite
results.

\subsection{Three distinct difficulties}

\emph{Nonstationarity} means that the relevant conditional laws or task
demands change over time. Time-indexed kernels can represent specified
changes, including recurring regimes. \emph{Nonergodicity} is a property
of a stationary process or invariant dynamical model: a trajectory need
not sample all invariant components according to the population law.
\emph{Path dependence} means that earlier actions and retention choices
affect the observations and revisions available later. An adaptive agent
may generate a nonstationary process, so the phrase ``nonergodic agent''
needs a specified invariant model before receiving a literal ergodic-theory
interpretation.

A stationary mixture with a persistent latent component can be
nonergodic even though its law does not change. A changing sequence of
regimes need not be a stationary mixture. A controlled trajectory can fail
to visit relevant situations because the agent never takes the revealing
action. SER accommodates those distinctions through the declared
conditional model and access restrictions. Context-dependent models and
lifelong abstractions already address parts of the surrounding problem
\citep{sodhani2022,abel2018}; historical repairability adds a further
constraint.

\begin{example}[A policy can hide a relevant distinction]
Two situations have the same outcome under action ``wait'' and different
outcomes under action ``use''. An agent that only waits has no observational
basis for concluding that the situations are equivalent under both actions.
A merge preserving the waiting policy may fail to preserve future
exploration. The protected continuation family must specify whether
``use'' is included, and any empirical certificate must respect the action
coverage actually available.
\end{example}

\section{Operational quotients and reusable dynamics}
\label{sec:operational}

\subsection{A quotient is a declared equivalence}

Let $X$ be a history or instance space, and let $\cF$ be a declared family
of response functions. Define
\begin{equation}
 x\sim_{\cF}x'
 \quad\Longleftrightarrow\quad
 f(x)=f(x')\ \text{for every }f\in\cF.
 \label{eq:operational}
\end{equation}
The functions can describe protected outputs, action-conditioned
predictions, or consequences of specified continuations. Their domain and
meaning must be fixed before equivalence is asserted. For a finite family,
the response profile $x\mapsto(f_1(x),\ldots,f_m(x))$ realizes the quotient.
Changing the response family can change the useful quotient.

Approximate pairwise similarity is not generally transitive. A relation
$|f(x)-f(x')|\le\epsilon$ does not by itself define quotient cells. An
approximate implementation can instead use a genuine partition with a
declared within-cell oscillation bound. That convention distinguishes
partition structure from an informal similarity graph.

\begin{proposition}[Classical quotient factorization]
\label{prop:descent}
Let $q:X\to Z$ be a quotient map of topological spaces and
$f:X\to[0,1]$ be continuous. A continuous map $\bar f:Z\to[0,1]$
satisfying $f=\bar f\circ q$ exists exactly when $f$ is constant on every
fiber of $q$. The descended map is unique. Whenever factorization exists
and $f(A)=\{0\}$, $f(C)=\{1\}$, the images $q(A)$ and $q(C)$ are
disjoint. Repeated factorization preserves the separator through a finite
quotient hierarchy.
\end{proposition}
\begin{proof}
If $q(x)=q(x')$ and $f=\bar f\circ q$, then $f(x)=f(x')$.
Conversely, define $\bar f(z)=f(x)$ for any $x\in q^{-1}(\{z\})$.
Surjectivity supplies a representative, and fiber constancy makes the value
well defined. For every open $V\subseteq[0,1]$,
\[
 q^{-1}(\bar f^{-1}(V))=f^{-1}(V).
\]
Continuity of $f$ and the defining property of a quotient topology show
that $\bar f^{-1}(V)$ is open. Surjectivity also gives uniqueness, since
any factorization must assign $\bar f(q(x))=f(x)$.

The quotient images lie in the disjoint sets
$\bar f^{-1}(\{0\})$ and $\bar f^{-1}(\{1\})$. For a hierarchy
$X_0\xrightarrow{q_1}\cdots\xrightarrow{q_D}X_D$, suppose recursively
that $f_{\ell-1}$ is constant on the fibers of $q_\ell$ and let
$f_{\ell-1}=f_\ell\circ q_\ell$, with $f_0=f$. Induction gives
$f=f_\ell\circ Q_\ell$ for $Q_\ell=q_\ell\circ\cdots\circ q_1$.
Thus $f_\ell$ takes value zero on $Q_\ell(A)$ and one on $Q_\ell(C)$
at every level, proving the claim.
\end{proof}

The disjointness conclusion is conditional on factorization. An arbitrary
quotient can identify points from $A$ and $C$. In a finite discrete space,
continuity is automatic, but fiber constancy remains a substantive
information condition.

\subsection{A separator-preserving ladder and local reopening}

Urysohn's lemma supplies a continuous $[0,1]$-valued separator for any
two disjoint closed sets in a normal space \citep{munkres2000}. Once
chosen, a separator survives a quotient precisely under the factorization
condition of Proposition~\ref{prop:descent}. We use \emph{Urysohn ladder}
as a descriptive name for a finite hierarchy whose declared separators
descend at the levels where their distinctions remain protected. The name
does not assert that every separator can survive every merge.

For nested partitions $P_0\preceq\cdots\preceq P_D$, let
$Q_\ell:X\to X/P_\ell$. A response $f$ is reusable at level $\ell$
when $f=g_\ell\circ Q_\ell$. A partition preserving a large
response family can support several coarser, cheaper interfaces for smaller
families. Generalization then comes from a response that is valid throughout
a cell, with its validity established under the declared model or evidence.
The existence of a cell alone supplies no evidence about unseen members.

Suppose a new target is constant within all old cells except one cell $C$.
A partition refinement confined to $C$ leaves every other address intact.
An existing response can be inherited by each child of $C$, preserving
protected competence while the new target receives a child-specific head.
That \emph{local reopening} is constructible only if the retained state
and admissible access recover the child address. Section~\ref{sec:repair}
states the exact information and migration conditions. A hierarchy offers
a place for repair; surviving evidence determines whether the repair is
available, and continuation value determines whether repair is worth its
cost.

\subsection{Preserving dynamics requires more than preserving one output}

For a finite fully observed controlled model, let $E$ be the state space,
$\cA$ the common action set, $P_t(e'\mid e,a)$ the time-indexed
transition kernel, $c_t(e,a)$ the stage cost, and $\psi(e)$ the terminal
cost. The following specialization of controlled state abstraction states
the additional requirements needed to preserve continuation value
\citep{ferns2004}.

\begin{theorem}[Value-preserving controlled quotient]
\label{thm:controlledquotient}
Let $q:E\to Z$ be surjective. Suppose that, for every $t,a,z,z'$ and
every $e$ with $q(e)=z$,
\begin{align}
 c_t(e,a)&=\bar c_t(z,a),\nonumber\\
 \sum_{e':q(e')=z'}P_t(e'\mid e,a)&=\bar P_t(z'\mid z,a),\label{eq:lump}\\
 \psi(e)&=\bar\psi(z).\nonumber
\end{align}
Then the optimal finite-horizon cost satisfies
$V_t(e)=\bar V_t(q(e))$. A minimizing abstract policy lifts to an optimal
policy in the original model.
\end{theorem}
\begin{proof}
At the terminal stage the conclusion follows from the last equality.
Assume $V_{t+1}(e')=\bar V_{t+1}(q(e'))$. For any action $a$,
\begin{align*}
 c_t(e,a)+\sum_{e'}P_t(e'\mid e,a)V_{t+1}(e')
 &=\bar c_t(q(e),a)
 +\sum_{z'}\bar P_t(z'\mid q(e),a)\bar V_{t+1}(z').
\end{align*}
The right-hand side depends on $e$ only through $q(e)$.
Minimizing over the common action set proves the induction step.
Choosing a minimizing action at each abstract state gives the lifted policy.
\end{proof}

The theorem assumes the action-conditioned equalities across all protected
states and actions, not merely those sampled by the current policy. A
partially observed learner does not automatically know or observe $q(E_t)$.
Section~\ref{sec:bellman} supplies a different planning state for that case.

\subsection{Conjugacy and operational transfer}

For deterministic systems with update maps $F_a$, a bijection $\phi$ of
states and a bijection $\beta$ of actions satisfying
\[
 \phi(F_a(e))=F'_{\beta(a)}(\phi(e))
\]
transports trajectories between systems. A topological conjugacy further
requires the appropriate topological regularity. To infer equal decision
value, costs, stopping criteria, access, and physical computation charges
must also transport correctly. A coordinate rescaling can preserve a phase
portrait while changing numerical conditioning, observation noise, or
hardware cost.

Operational transfer can be weaker than conjugacy. Theorem~\ref{thm:controlledquotient}
allows multiple states to share one abstract state when their relevant
costs and abstract transitions agree. The resulting quotient is an
operational index of reusable dynamics. Calling that index a moduli space
does not establish that the index has a manifold structure or a useful
metric. Any geometric realization needs its own assumptions.

For learning-to-optimize, a problem instance and the current optimizer state
can together play the role of $e$, while $a$ chooses an update rule.
Reusable optimization mechanisms require equivalence of the relevant
progress, stopping, and resource responses. A common qualitative phase
portrait alone need not preserve convergence speed or implementation cost.
Bellman selection can then compare acquiring a new mechanism, reusing a
stored mechanism, or repairing a mechanism under changed demands.

\section{Split, merge, and historical repair}
\label{sec:repair}

\subsection{Active organization and complete retained information}

Fix a finite historical record $X$ with support $\cX$. A deployed
coordinate is $Z=q(X)$. The complete retained state is $S=r(X)$ and may
include an auxiliary witness buffer, representation descriptors, learned
parameters, routers, and audit records. The coordinate $q(X)$ is
recoverable from $S$. Every record-dependent choice of architecture,
allocation, or codebook belongs to the information-bearing retained state.
Fixed, record-independent programs also incur their declared storage and
computation charges.

A pure coordinate merge has the form $q^-=a\circ q$. A pure split is a
refinement $q^+$ satisfying $q=a\circ q^+$. The parent map $a$ identifies
which old address each child belongs to. General representation edits need
not be pure splits or merges.

\begin{proposition}[Protected merge and inherited split]
\label{prop:splitmerge}
Let the protected responses satisfy $f_j=g_j\circ q$.
A merge $q^-$ preserves those responses exactly if and only if each
$f_j$ is constant on every fiber of $q^-$. For a refinement
$q=a\circ q^+$, assigning $g_j^+=g_j\circ a$ preserves every protected
response exactly.
\end{proposition}
\begin{proof}
The merge statement follows from set-theoretic factorization through a
surjection, or from Proposition~\ref{prop:descent} when continuity is
required. For the split,
$g_j^+(q^+(x))=g_j(a(q^+(x)))=g_j(q(x))=f_j(x)$.
\end{proof}

Inheritance preserves outputs only under the corresponding actual routing.
Introducing a child router can change behavior if that router sends an
input to a different parent. Routers and copied heads also consume
resources. The proposition establishes output preservation, not a free
implementation or a generalization guarantee for unrelated targets.

Merging active interfaces and erasing witnesses are separate decisions.
A shared interface can replace several active computations while an
auxiliary buffer preserves their historical distinction. Conversely, a
coordinate-only merge can preserve current outputs while removing every
route to a later useful repair.

\subsection{The exact information budget for a specified replacement}

Let $Y=q^+(X)$ be a specified replacement coordinate, and let future access
$W$ be finite valued with a known joint law with $X$. An encoder forms
$M=e(X)\in\{0,1\}^b$ before observing $W$. A decoder uses $(Z,M,W)$.
For every nonempty fiber $\cX_z=q^{-1}(z)\cap\cX$, form a graph $G_z$
whose distinct vertices $x,x'$ are adjacent when
\begin{equation}
 q^+(x)\ne q^+(x')
 \quad\text{and}\quad
 \supp P_{W\mid x}\cap\supp P_{W\mid x'}\ne\varnothing.
 \label{eq:edges}
\end{equation}
Such histories require different repairs and can yield the same future
access. All graph and fiber statements are restricted to positive-probability
histories.

\begin{theorem}[Fixed-length revision evidence]
\label{thm:color}
The minimum auxiliary budget permitting
$\Prob(\widehat Y=Y)=1$ is
\begin{equation}
 b^*(q,q^+;W)=\left\lceil\log_2\max_z\chi(G_z)\right\rceil.
 \label{eq:color}
\end{equation}
Here $\chi=1$ for a nonempty edgeless graph. If $W$ is constant,
\begin{equation}
 b^*(q,q^+;W)=\left\lceil\log_2\max_z K_z\right\rceil,
 \qquad K_z=|q^+(\cX_z)|.
 \label{eq:count}
\end{equation}
\end{theorem}
\begin{proof}
For any zero-error code, the decoder is correct on every supported pair
$(x,w)$ because the model is finite. If adjacent histories shared a message,
their common $z$, message, and an overlapping observation $w$ would give
identical decoder inputs but different required outputs. Thus $e$ is a
proper coloring on each $G_z$ with at most $2^b$ colors, proving the
lower bound.

Set $m=\max_z\chi(G_z)$ and choose proper colorings
$c_z:\cX_z\to\{1,\ldots,m\}$. Encode the color with
$b=\lceil\log_2m\rceil$ bits, reusing the color labels across fibers
because the decoder observes $z$. For a decoder input $(z,c,w)$ define
\[
 \cC(z,c,w)=\{x\in\cX_z:c_z(x)=c, P(W=w\mid X=x)>0\}.
\]
If nonempty, the set has one common replacement value. Two members with
different replacement values would be adjacent by their shared $w$, in
conflict with their common color. Decode that common value, assigning an
arbitrary replacement value to an empty set. On every supported input the
true history belongs to the compatible set, so decoding is exact.

If $W$ is constant, $G_z$ is complete multipartite, with one nonempty
part per replacement value. Coloring by replacement uses $K_z$ colors.
Selecting one vertex from each part gives a clique of size $K_z$.
Hence $\chi(G_z)=K_z$, proving the second formula.
\end{proof}

The result is a finite one-shot specialization of classical zero-error
side-information and functional coding \citep{witsenhausen1976,orlitsky2001}.
The optimization chooses a code while the record remains available. The result does not
imply that an arbitrary deployed memory of the same size can be converted
into an attaining code. Independent code randomization cannot lower the
zero-error budget: fixing all independent random seeds to a realization
with zero conditional error yields a deterministic code with the same
message budget. The finite support and probability-one requirement ensure
that such realizations exist.

\subsection{The actual memory determines which repairs are reachable}

\begin{proposition}[Causal migration]
\label{prop:migration}
For complete current and replacement states $S=r(X)$ and $S'=r'(X)$,
an exact update $S'=M(S,W)$ exists on the finite support if and only if
\begin{equation}
 r(x)=r(x'),\quad
 \supp P_{W\mid x}\cap\supp P_{W\mid x'}\ne\varnothing
 \quad\Longrightarrow\quad r'(x)=r'(x').
 \label{eq:migration}
\end{equation}
\end{proposition}
\begin{proof}
Necessity follows because a common supported pair $(s,w)$ must have one
update output. For sufficiency, assign $M(s,w)$ the value $r'(x)$ for
any supported $x$ with $r(x)=s$ and $P(W=w\mid x)>0$.
The condition makes the assignment independent of the representative.
Assign arbitrary values on unsupported pairs.
\end{proof}

For a fixed future-access law, define the feasible replacement family
\begin{equation}
 \cR(r;W)=\{r':r'(X)=M(r(X),W)\text{ almost surely for some }M\}.
 \label{eq:reachable}
\end{equation}
The update can replace both coordinate and evidence. A coordinate-only
repair is the special case $r'=q^+$.

\begin{corollary}[Erasure can only remove fixed-access repair options]
\label{cor:erasure}
If $r_2=g\circ r_1$, then
$\cR(r_2;W)\subseteq\cR(r_1;W)$.
\end{corollary}
\begin{proof}
Any update $M(r_2(X),W)$ is also the update $M(g(r_1(X)),W)$.
\end{proof}
The comparison fixes the same access law. If a different memory changes
which actions and observations become available, the joint controlled
model must be compared instead.

\begin{example}[One merge, two different futures]
\label{ex:bits}
Let $X=(U,V)$ contain independent fair bits and protect response $U$.
Merge the four active $(U,V)$ cells into two $U$ cells. Retaining
$S_1=(U,B)$ with $B=V$ preserves the split back to $(U,V)$ under constant
future access. Retaining $S_0=U$ does not. Both active organizations
support the same current output, but their repair families differ.
The buffer costs one bit, so $S_1$ still contains two historical bits.
Any advantage of the active merge must come from a declared saving in
interface, computation, or maintenance cost.
\end{example}

\subsection{How the objective differs from split--merge VQ}

Classical VQ optimizes reconstruction distortion under a codebook or rate
constraint. General distortion measures are allowed \citep{linde1980}.
Split--merge codebook methods reallocate codevectors when distortion
reduction from splitting exceeds distortion increase from merging
\citep{kaukoranta1998}. Their practical operations can include global
reassignment, so the resulting partitions need not be nested.

\begin{table}[t]
\centering\small
\caption{The distinction is in the objective and admissibility conditions,
not in the split and merge primitives.}
\label{tab:vq}
\begin{tabularx}{\linewidth}{@{}p{.18\linewidth}XX@{}}
\toprule
Question & Conventional distortion-based VQ & SER retention and consolidation\\
\midrule
What is valued? & Reconstruction under a specified distortion and rate model.
& Competence and supported continuation value under resource costs.\\
Why split? & Improve the chosen distortion objective.
& Enable a supported distinction whose repair is causally implementable.\\
Why merge? & Release or reallocate representational resolution.
& Share protected computation while deciding separately what evidence survives.\\
What limits repair? & The algorithm's data, model, and coding constraints.
& The actual retained state and admissible future access.\\
When install? & The method's optimization or update rule.
& A declared prospective evidence criterion and migration check.\\
\bottomrule
\end{tabularx}
\end{table}

VQ can serve as an implementation inside an SER controller. Adding
protected-response constraints, witness retention, access restrictions, and
continuation value changes the optimization problem. A contrast between
``geometry'' and ``semantics'' would miss that distinction: a VQ distortion
can already be task dependent, and an SER realization can use metric
quantization.

\section{Recurrence, nonergodicity, and the value of retention}
\label{sec:recurrence}

Recurrence concerns the return of a situation or demand. Reuse further
requires that a retained mechanism remains applicable and can be addressed
correctly. Repair further requires the historical distinctions needed to
construct an update. A return of the same task type does not necessarily
reveal the missing content of a past episode.

\subsection{The classical recurrence theorem has a limited role}

\begin{theorem}[Poincar\'e recurrence, classical]
\label{thm:poincare}
Let $T$ preserve a finite measure $\mu$ on a measurable space. For every
measurable set $A$, almost every point of $A$ returns to $A$ infinitely
often under iteration of $T$. Ergodicity is not required.
\end{theorem}
\begin{proof}
Let $B\subseteq A$ be the measurable set of points that never return to
$A$ after time zero. The sets $T^{-n}B$, $n\ge0$, are pairwise disjoint.
Indeed, membership in $T^{-n}B$ and $T^{-m}B$ with $n<m$ would give a
point of $B$ that returns to $B\subseteq A$ after $m-n$ steps.
Measure preservation gives $\mu(T^{-n}B)=\mu(B)$, so finiteness of the
ambient measure forces $\mu(B)=0$. Any point of $A$ with finitely many
returns has a last visit, whose image belongs to $B$. Such points lie in
$\bigcup_{n\ge0}T^{-n}B$, a null set.
\end{proof}

The result gives no useful finite-horizon return bound by itself. In
probability-preserving invertible systems, the generalized return-time
identity is $\int_A\tau_A\,d\mu=\mu(I_A)$, where $I_A$ is the invariant
saturation of $A$; under ergodicity and $\mu(A)>0$, the conditional mean
return time becomes $1/\mu(A)$ \citep{baezduarte2005}. Those conclusions
require their respective hypotheses. Stationary workload generation alone
does not imply that an adaptive agent's state update preserves a fixed
finite measure.

For a nonergodic invariant model, the return law can depend on the
component occupied by the trajectory. A population frequency need not be
the relevant frequency for the current agent. For a nonstationary or
controlled model, we specify return hazards directly or infer them under
an explicit predictive model.

\subsection{A finite first-return allocation problem}

Suppose one historical witness $V$ is initially available. Protected
competence is unaffected by retaining or erasing $V$. A single revision
opportunity arrives at its first return time $\tau$, or never arrives
within a horizon $T$. Conditional on survival to stage $t$, the specified
hazard is
\[
 h_t=\Prob(\tau=t\mid\tau\ge t).
\]
The return process and its audit supply no information about $V$, and
erasure prevents exact installation permanently within the horizon. The
opportunity's declared evidence criterion has already been met; its
conditional expected net installation gain is $g_t\ge0$, independent of
the actual witness value. All candidate-specific edit and audit charges
are included in $g_t$. Keeping the witness through stage $t$ incurs
$\lambda_t\ge0$, including the stage in which a return occurs.
Forgetting is free in the benchmark, and future values are discounted by
$\delta\in(0,1]$.

The information available before a return consists only of the stage and
the fact that no return has occurred. A more informative context would
require a conditional hazard state in the recursion below. Restricting
pre-return information to stage and survival makes the scalar calculation
exact.

\begin{theorem}[First-return retention value]
\label{thm:hazard}
Let $R_t$ be the maximum incremental expected value of possessing the
witness at stage $t$, relative to prior erasure, conditional on no
earlier return. Then
\begin{equation}
 R_{T+1}=0,\qquad
 R_t=\max\{0,-\lambda_t+h_tg_t+(1-h_t)\delta R_{t+1}\}.
 \label{eq:hazard}
\end{equation}
The witness is retained through stage $t$ exactly when the second
term is positive, with either action optimal at equality.
\end{theorem}
\begin{proof}
Erasure yields zero incremental continuation value. Retention incurs
$\lambda_t$. A return occurs with probability $h_t$ and contributes
$g_t$, after which the single-opportunity problem ends. Otherwise the
witness remains available for the next stage, whose optimal value is
$\delta R_{t+1}$. These are all supported possibilities. Backward
induction gives the recursion and an attaining retain-or-erase policy.
\end{proof}

For the particular strategy that always retains until the first return
or the horizon, the expected value is
\begin{equation}
 \E\left[
 \sum_{t=1}^T\delta^{t-1}g_t\ind\{\tau=t\}
 -\sum_{t=1}^{\min\{\tau,T\}}\delta^{t-1}\lambda_t
 \right],
 \label{eq:returnvalue}
\end{equation}
where $\min\{\infty,T\}=T$. The optimal value can be larger because
the witness may be abandoned before the horizon.

\begin{example}[Immediate value can miss a delayed repair]
\label{ex:delayed}
Let $T=3$, $\delta=1$, $h_1=h_2=0$, $h_3=1$, $g_3=2$, and
$\lambda_t=1/5$ at every stage. The optimal value is $R_1=7/5$.
A myopic rule comparing $h_tg_t$ against $\lambda_t$ erases the witness
at the first stage and achieves zero. The gain is an exact consequence of
the specified return and access model, not an empirical result.
\end{example}

\begin{proposition}[Guaranteed recurrence need not justify retention]
\label{prop:longreturn}
For a deterministic return at time $L$, fixed $g>0$, maintenance
$\lambda>0$, and $0<\delta<1$, retaining until return has value
\begin{equation}
 \delta^{L-1}g-\lambda\frac{1-\delta^L}{1-\delta}.
 \label{eq:longreturn}
\end{equation}
The value is negative for sufficiently large $L$. For $\delta=1$,
the value is $g-L\lambda$, with the same conclusion.
\end{proposition}
\begin{proof}
Sum the discounted maintenance charges and the single discounted gain.
For $\delta<1$, the gain tends to zero and the maintenance sum tends to
$\lambda/(1-\delta)>0$. The undiscounted formula is immediate.
\end{proof}

A finite cyclic measure-preserving system can realize an arbitrarily long
return period. Thus recurrence does not determine an economically optimal
memory policy. Bellman continuation value depends on the timing of
recurrence, witness availability, useful gain, and resource costs.

The scalar theorem does not imply that independently pricing witnesses
solves allocation under a shared capacity constraint. Competing witnesses,
jointly required repairs, information acquisition, and action-dependent
return laws can couple the decisions. Their allocation belongs to the
joint problem of Section~\ref{sec:bellman}.

\section{Bellman control over representations and retained information}
\label{sec:bellman}

\subsection{The learner's causal interface}

Fix a horizon $T$ and finite alphabets for hidden state $E_t$, deployed
memory $S_t$, observation $O_t$, physical action $A_t$, and feedback $F_t$.
The hidden state includes relevant inaccessible history, latent regimes,
and any variables needed to make the declared transition model Markov.
A static historical record can be carried as a hidden component of $E_t$.
Time-indexed kernels represent specified nonstationarity:
\[
 O_t(o\mid e),\qquad P_t(e',f\mid e,o,a).
\]
The learner first observes $(S_t,O_t)$, chooses an action, receives
feedback, and updates memory:
\begin{equation}
 A_t=\pi_t(S_t,O_t),\qquad
 S_{t+1}=u_t(S_t,O_t,A_t,F_t).
 \label{eq:causal}
\end{equation}
Splits, merges, witness allocation, routing edits, and evidence acquisition
are encoded in those causal rules and the controlled feedback law.

Fix finite rule catalogs and a computational accounting convention. A
common record-independent program allocation can be charged once.
Alternatively, installed program and representation descriptors can be
included in $S_t$, with a fixed interpreter restricting executable updates.
Record-dependent choices cannot reside in an uncharged external policy
table. These conventions make the finite optimization an implementation
benchmark with explicit information charges.

The expected lifetime objective is
\begin{equation}
 J_T= C_{\rm prog}+
 \E\left[\sum_{t=1}^T
 \bigl(\ell_t+\lambda_M M_t+\lambda_C C_t+
       \lambda_D D_t\bigr)+\Psi(E_{T+1},S_{T+1})\right].
 \label{eq:lifetime}
\end{equation}
Here $\ell_t$ is task loss, $M_t$ complete memory use, $C_t$ computation,
and $D_t$ structural edit cost, with stated nonnegative weights.
Evidence acquisition incurs loss or resource cost through the same model.
The symbol $D_t$ here denotes edit cost; paired audit differences
are written $L_{a,i}$ in Section~\ref{sec:commitment}.

Protected competence, fixed memory budgets, exact migration, and a declared
confidence gate define admissibility. Constraints that depend on additional
history require corresponding controller or planning variables. A
nonadditive program-description cost cannot be silently replaced by a
sum of stage costs.

\subsection{Why the compressed coordinate is not always a Bellman state}

Present behavior can be identical under two memories with different repair
families, as in Example~\ref{ex:bits}. More basically, two histories that
produce the same memory value can imply different future environment laws.
An arbitrary compressed coordinate need not be a sufficient information
state. Computing a posterior from discarded observations and supplying that
posterior to the learner would undo the memory restriction.

For an offline designer, use the joint law before the current observation,
\begin{equation}
 \eta_t(e,s)=\Prob(E_t=e,S_t=s).
 \label{eq:eta}
\end{equation}
At stage $t$, a \emph{prescription} $\gamma=(\pi,u)$ is a pair of complete
causal rules. The designer chooses rules; the deployed learner applies
those rules to the actual retained state and observations. The distribution
$\eta_t$ is not a deployed memory register.

The prescription induces the deterministic joint-law update
\begin{align}
 \mathcal F_t(\eta,\gamma)(e',s')
 ={}&\sum_{e,s,o,f}\eta(e,s)O_t(o\mid e)
 P_t(e',f\mid e,o,\pi(s,o))\nonumber\\
 &\hspace{6mm}\times
 \ind\{s'=u(s,o,\pi(s,o),f)\}.
 \label{eq:eta-update}
\end{align}
The randomness of observations and transitions is already integrated into
the updated distribution. That is why the Bellman equation below has no
additional expectation outside its continuation value.

Let $\cG_t(\eta)$ be the finite set of admissible prescriptions. Its
definition can impose support-based migration, competence preservation,
resource limits, and a fixed gate implemented in charged audit registers.
A lifetime chance constraint needs a remaining-risk variable or a terminal
constraint on an augmented failure state unless a fixed gate already
enforces the constraint. All such state is included explicitly.
Let $\bar c_t(\eta,\gamma)$ be the expected stage cost under the finite
law, including a common program charge at the initial stage when needed.

\begin{theorem}[Finite SER prescription principle]
\label{thm:bellman}
Under the preceding finite catalogs, additive accounting, and sufficient
augmented-state conventions, the minimum expected cost is given by
\begin{align}
 V_{T+1}(\eta)&=\sum_{e,s}\eta(e,s)\Psi(e,s),\nonumber\\
 V_t(\eta)&=\min_{\gamma\in\cG_t(\eta)}
 \{\bar c_t(\eta,\gamma)
       +V_{t+1}(\mathcal F_t(\eta,\gamma))\}.
 \label{eq:bellman}
\end{align}
An empty admissible set has value $+\infty$. Whenever a feasible policy
exists, backward induction attains the optimum over every representation
and evidence trajectory generated by the declared catalogs.
\end{theorem}
\begin{proof}
Given earlier prescription choices, $\eta$ specifies the complete current
joint law. For any first prescription $\gamma$, the expected stage cost
is $\bar c_t(\eta,\gamma)$ and the next law is
$\mathcal F_t(\eta,\gamma)$. Every admissible strategy has such a first
prescription followed by an admissible continuation. By induction, its
tail cost is at least $V_{t+1}$ at the induced law, giving the lower
bound in the recursion. Conversely, choose a minimizing first prescription
and an attaining optimal continuation at the induced law. The finite rule
catalog and finite horizon supply a minimizer whenever the subproblem is
feasible. Induction from the terminal cost proves equality and attainment.
\end{proof}

The theorem is an application of Bellman's principle, with the control
space enlarged to include retained information and structural updates.
Distribution-state and prescription methods have precedents in stochastic
control \citep{nayyar2013,tottori2022}. The SER specification contributes
the particular migration, competence, access, and evidential restrictions
within that optimization problem.

The probability simplex is continuous, but only finitely many joint laws
are reached from a fixed initial law by finite prescription sequences over
a finite horizon. Their number can still grow exponentially. The theorem
is an exact characterization with potentially expensive computation. The recursion optimizes
deterministic rules in the declared catalogs. Randomized policies can
require an expanded catalog, especially when feasibility is imposed only
through expectation or chance constraints.

\subsection{Memory has continuation value}

For a feasible prescription, define
\[
 Q_t(\eta,\gamma)=\bar c_t(\eta,\gamma)
      +V_{t+1}(\mathcal F_t(\eta,\gamma)).
\]
A structural edit is preferred over reuse when its Bellman cost is smaller.
If the edit incurs an additional immediate charge $\kappa$, improvement
in the remaining terms must exceed $\kappa$. The comparison includes
future routing, acquisition, witness retention, repair, and retirement.
Reducing present prediction error alone does not characterize the optimum.

The continuation benefit of a witness is a difference between the values
of the induced future information states. The difference is generally
context dependent and nonadditive across witnesses. Capacity constraints
and repairs requiring several witnesses can create interactions that a
single scalar relevance score misses.

\begin{example}[One-step allocation]
Before historical deletion, choose either to retain a witness for total
cost $c$ or erase the witness. A later independent audit identifies one usable
opportunity with probability $p$, certifies conditional expected net gain
$G>0$, and reveals no missing witness information. Exact installation
requires the witness. With discount $\delta$, retention is optimal when
$\delta pG>c$, erasure is optimal when the inequality is reversed, and
equality gives a tie. The result is the two-stage Bellman comparison under
the stated access and gain assumptions.
\end{example}

\subsection{Global paths and the finite-memory limit}

Early retention decisions determine later feasible updates. The optimizer
compares complete paths of representations and evidence. At the
deployment level, each continuation remains restricted to the information
and executable programs that actually survived.

Finite-horizon induction requires no ergodicity or stationarity assumption.
The derivation requires specified conditional kernels and a sufficient planning state.
For unknown dynamics, the model must include the inference state, or the
objective must specify a robust uncertainty set. No model-free solution
to arbitrary unannounced change follows from the recursion.

\begin{proposition}[Independent historical demands require capacity]
\label{prop:independentbits}
Suppose $n$ independent fair historical bits are all potentially requested,
the decoder learns which bit is requested, and future access reveals none
of their values. Exact answering of every possible request requires at
least $n$ retained bits.
\end{proposition}
\begin{proof}
If two distinct bit strings shared a retained codeword, they would differ
in some requested bit. The decoder would receive the same codeword and
request but need different outputs. Thus the code must distinguish all
$2^n$ strings and needs at least $n$ bits.
\end{proof}

Split--merge can exploit redundancy and selective relevance. Preserving
an unbounded number of independent, permanently relevant historical
distinctions requires increasing memory capacity.

\section{When additional dimensions support learning}
\label{sec:dimension}

Ambient dimension, the number of relevant factors, interaction order, and
the number of informative observation channels are different quantities.
SER does not automatically convert high dimension into a benefit. Two
explicit models show where favorable scaling can arise and what the
benefit requires.

\subsection{Factor sharing avoids storing every combination}

Let $z=(z_1,\ldots,z_m)\in\{1,\ldots,K\}^m$. A flat table has
$K^m$ entries. Suppose a protected response admits the factorization
\begin{equation}
 f(z)=\sum_{J\in\cI}f_J(z_J),\qquad |J|\le k,
 \label{eq:factor}
\end{equation}
where $\cI$ is a specified family of factor subsets. Assume known factor
addresses and a fixed summation implementation.

\begin{proposition}[Conditional storage and local repair bound]
\label{prop:factor}
If every table value uses $p$ bits, the response in
\eqref{eq:factor} can be represented with
\begin{equation}
 p\sum_{J\in\cI}K^{|J|}+C_{\rm route}+C_{\rm meta}
 \le p|\cI|K^k+C_{\rm route}+C_{\rm meta}
 \label{eq:factorcost}
\end{equation}
bits, where routing and metadata costs are explicitly included.
If only terms in $\cI_{\rm edit}\subseteq\cI$ change, replacement data
uses at most $p\sum_{J\in\cI_{\rm edit}}K^{|J|}$ bits, apart from
its routing, version, audit, and edit overhead. Evaluation needs one lookup
per term and the corresponding additions.
\end{proposition}
\begin{proof}
Store each table $f_J$ with $K^{|J|}$ entries and evaluate the declared
sum. Updating a term leaves all other stored tables and their contributions
unchanged. Count the replaced entries and include the remaining declared
overhead separately.
\end{proof}

For additive responses, $\cI=\{\{1\},\ldots,\{m\}\}$, so the
table data uses $pmK$ bits and covers $K^m$ possible combinations.
Increasing the number of factors expands the supported combination space
much faster than the stored table count. If a new demand introduces a
sparse interaction, an SER controller can propose a local interaction
term, retain the witnesses needed for construction, and audit the benefit.
That proposal does not establish that the term is discoverable from the
available data or that unrestricted global planning becomes tractable.

The fixed precision assumption concerns exact storage of the declared
table values. For approximate real-valued values, entry error $\epsilon$
can accumulate to $|\cI|\epsilon$ in the sum. A total error target
requires suitable per-entry precision, which can increase $p$. Factor
discovery and maintenance costs also belong to the objective.

Factored MDPs and structured value functions provide related complexity
reductions under explicit assumptions \citep{deng2022}. A nested ladder
alone need not supply the sharing in \eqref{eq:factor}: a hierarchy can
still have exponentially many leaves. Multiple views or shared modules
can be economical for cross-cutting demands. For example, partitions by
$z_1$ and by $z_2$ are generally incomparable under refinement, although
both descend from a common refinement.

\begin{proposition}[No compression of arbitrary response tables]
\label{prop:tablelower}
Representing every binary response function on $K^m$ combinations
exactly requires at least $K^m$ bits in the worst case, including any
function-dependent program or architecture description.
\end{proposition}
\begin{proof}
There are $2^{K^m}$ binary functions. Distinct functions must have distinct
complete encodings because they differ on at least one query. A fixed
budget $b$ can distinguish at most $2^b$ encodings, so $b\ge K^m$.
\end{proof}

\subsection{Redundant informative channels improve identification}

Let $V$ be a persistent fair regime bit and observe $D$ channels
$W_i=V\oplus N_i$, where the $N_i$ are independent Bernoulli noise bits
with $0<\rho<1/2$. The independence is conditional on $V$ and the bit
remains fixed across all channels. Let $\widehat V$ be majority voting,
with any declared rule for ties.

\begin{proposition}[Approximate identification improves while exact budget stays fixed]
\label{prop:channels}
The majority error satisfies
\begin{equation}
 \Prob(\widehat V\ne V)
 \le\exp\{-2D(1/2-\rho)^2\}.
 \label{eq:majority}
\end{equation}
If a deployed coordinate is independent of the pair $(V,W_{1:D})$, and exact historical
revision requires $V$, the auxiliary zero-error budget remains one bit
for every finite $D$.
\end{proposition}
\begin{proof}
Conditional on either value of $V$, an error requires
$\sum_iN_i\ge D/2$, including a possible tie. Hoeffding's inequality
gives the displayed bound \citep{hoeffding1963}.
For exact recovery, every observation vector has positive probability
under both values of $V$ because $0<\rho<1$. Independence of the deployed
coordinate from the joint pair prevents the coordinate from revealing
the channel noise. The two historical values
remain adjacent in the revision graph. One bit is necessary and retaining
$V$ attains the requirement.
\end{proof}

Additional informative channels can strengthen routing or audit evidence.
Perfectly correlated copies do not multiply independent evidence, and
irrelevant dimensions provide no such guarantee. The cost of sensing,
processing, and storing statistics also grows unless an implementation
provides a compensating reduction elsewhere.

The channel model concerns a persistent latent bit. If each new episode
draws an independent bit, observations of later episodes do not identify
an erased old bit. Recurrence of a situation type and recovery of an
episodic witness must be kept separate.

The two propositions distinguish avoiding exhaustive combinations from
gaining statistical evidence. A favorable SER architecture exploits factor
reuse while preserving selected exceptions, and uses informative channels
without paying to store every raw observation indefinitely. The Bellman
objective decides whether those benefits exceed all associated costs.

\section{A concrete evidential commitment constraint}
\label{sec:commitment}

Knowing that an edit is representable and causally constructible does not
establish that installation will improve future performance. The following
finite gate supplies one admissible commitment rule and a sufficient
certificate. An optimal stopping rule would require a separate derivation.

Freeze a baseline and at most $K$ candidates before collecting a fresh
passive audit. For candidate $a$, let $L_{a,i}\in[-1,1]$ be the paired
baseline-minus-candidate loss and
$\mu_{a,i}=\E[L_{a,i}\mid\cF_{i-1}]$ its predictable mean.
Potential paired losses must be defined on the same observation process
across audit and deployment. Installing the candidate must preserve the
observation law used in the comparison. An action-dependent change of
that law requires separate counterfactual or off-policy justification.

Fix an audit cap $N$, prospective horizon $H$, risk $\alpha\in(0,1)$,
and candidate charge $\kappa_a\ge0$. Assume a supplied pathwise drift
envelope $|\mu_{a,r}-\mu_{a,s}|\le\nu|r-s|$ over the audit and
prospective indices. At any $n\le N$, define
\begin{equation}
 \mathcal L_a(n)=\frac1n\sum_{i=1}^nL_{a,i}
 -\sqrt{\frac{2\log(KN/\alpha)}n}
 -\frac{\nu(n+H)}2-\frac{\kappa_a}H.
 \label{eq:gate}
\end{equation}
Commit only if migration and protected competence are feasible and the
chosen candidate has $\mathcal L_a(n)>0$.

\begin{theorem}[Finite supported commitment]
\label{thm:gate}
Under the stated assumptions, with probability at least $1-\alpha$,
every authorized candidate and inspection time has positive conditional
expected net gain over its declared horizon.
\end{theorem}
\begin{proof}
For a fixed candidate and $n$, the centered differences have conditional
range length at most two. The conditional exponential bound gives
\[
 \Prob\left\{\frac1n\sum_{i=1}^n(L_{a,i}-\mu_{a,i})
 >\sqrt{2\log(KN/\alpha)/n}\right\}\le\frac\alpha{KN}.
\]
A union bound over $K$ candidates and $N$ times gives a simultaneous
event of probability at least $1-\alpha$. Bounded conditional differences
permit dependence across time \citep{azuma1967}.
On that event, use the drift envelope and take conditional expectations
at the inspection time to obtain
\begin{align*}
 \frac1H\sum_{h=1}^H\E[\mu_{a,n+h}\mid\cF_n]
 &\ge\frac1n\sum_{i=1}^n\mu_{a,i}
 -\frac\nu{nH}\sum_{i=1}^n\sum_{h=1}^H(n+h-i)\\
 &=\frac1n\sum_{i=1}^n\mu_{a,i}-\frac{\nu(n+H)}2.
\end{align*}
The conditional expected net gain is at least $H\mathcal L_a(n)$.
Positive gate values certify the claim simultaneously.
\end{proof}

For repeated rounds starting at stopping times, allocate deterministic
$\alpha_j>0$ with $\sum_j\alpha_j\le\alpha$ and restart with frozen
candidates and fresh audit observations. Conditioning on each round's
start and taking a countable union bound gives the same lifetime failure
budget. Audit sums, counters, candidate versions, and risk accounting
consume memory and enter the deployed state.

The certificate needs a supplied drift envelope and a finite commitment
horizon. Arbitrary future changes defeat an unrestricted prospective
guarantee. Within the finite Bellman problem, the gate restricts the allowed
prescriptions and its delay and resource costs are included in the
objective. A globally optimal gated controller is optimal among controllers
obeying that gate, not among all possible commitment rules.

\section{Finite verification and analytical illustrations}
\label{sec:checks}

The accompanying package contains a deterministic verification script and
generated numerical records. The checks are finite enumerations of
specified models and analytical illustrations. They are not measurements
from a trained agent and do not establish biological or open-ended learning
claims. The random seed is \texttt{20261001}.

\subsection{Independent checks of the finite calculations}

\paragraph{Coding.}
For each of 48 random six-history models, the script draws a deployed
coordinate with two or three possible labels, a replacement with three
possible labels, and a nonempty subset of three access symbols for each
history. The histories have positive uniform probabilities. Conditional
access probabilities can be chosen uniform on each specified subset; only
their supports affect zero-error feasibility. The same coordinate and
replacement are also checked with constant access, giving
\CodingModels{} models in total.

All \EncoderPartitions{} set partitions of the six histories are treated
as possible deterministic message encodings, up to message-label
permutation. For each encoding, decoder feasibility is checked directly:
every supported tuple $(z,m,w)$ must require one replacement value.
The smallest feasible number of message classes is compared with the
maximum fiber chromatic number computed by a separate coloring search.
After \DecoderChecks{} decoder-partition checks, the exhaustive optimum
agrees with Theorem~\ref{thm:color} in all \CodingModels{} models. The
constant-access cases also match the distinct-replacement count.

\paragraph{First-return allocation.}
For \HazardModels{} seeded models with horizons from one to nine stages,
hazards in $\{0,1/4,1/2,3/4,1\}$, nonnegative gains and maintenance
charges, and discounts in $\{3/4,1\}$, the recursion is evaluated using
exact rational arithmetic. Every deterministic policy before the first
return is equivalent to a cutoff: retain through a chosen number of stages,
then erase if no return has occurred. Exhaustive evaluation of all
\HazardPolicies{} cutoff policies agrees with the recursion in every
model. Example~\ref{ex:delayed} separately returns the exact value $7/5$.

\paragraph{Prescription optimization.}
A three-stage partially observed model has a persistent fair hidden bit,
a memory register in $\{0,1,\bot\}$, and observation error probabilities
$(1/5,9/20,1/10)$. Initially the register is empty. Four action rules use
the register when available, use the observation, always predict zero, or
always predict one. Four update rules retain or erase the register, write
the observation, or write the selected action. Their 16 combinations form
the per-stage catalog. The stage cost is prediction error plus a register-use
charge $2/25$ whenever the updated register is nonempty and an edit charge
$3/100$ whenever the register changes. The common interpreter and rule
catalog are fixed and contribute the same omitted additive charge to every
sequence; the variable register charges are benchmark cost units, not a
claim about hardware bit counts.

Joint-law backward induction agrees with exhaustive evaluation of all
\PrescriptionSequences{} three-stage prescription sequences, yielding
cost $\PrescriptionValue$. The computation verifies the declared causal
interface and deterministic-law recursion on a small instance. Exhaustive
search selects rules before deployment; the learner receives only its
actual register and observation. No hidden-state posterior is supplied as
uncharged memory.

\paragraph{Dimensions and local repair.}
Exact binomial tail probabilities obey the channel bound in all
\ChannelModels{} tested combinations of odd $D\le31$ and
$\rho\in\{1/20,2/20,\ldots,9/20\}$. In \FactorModels{} factor-table
models with $m=2,\ldots,6$ and $K\in\{2,3\}$, the script constructs
additive responses with one pair interaction. The local repair replaces the pair table
and verifies that the local difference update equals complete reevaluation
at all \FactorQueries{} input combinations. The constructed response
families illustrate the storage and locality conditions; they do not test
discovery of an unknown factorization.

\begin{figure}[t]
\centering
\includegraphics[width=\linewidth]{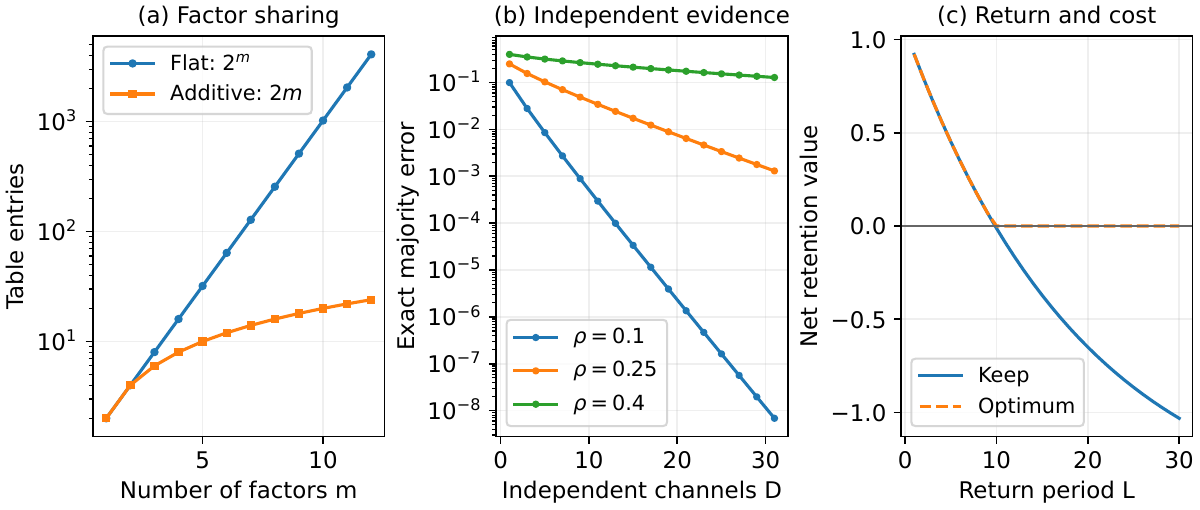}
\caption{Analytical models, not learned-agent experiments. (a) A binary
additive response uses $2m$ table entries to cover $2^m$ combinations;
routing, precision, and metadata costs remain additional. (b) Exact
binomial majority errors for independent noisy channels observing one
persistent bit. The exact auxiliary budget stays one bit at every finite
$D$ under Proposition~\ref{prop:channels}. (c) Retention until a
deterministic return with $g=1$, $\lambda=0.08$, and $\delta=0.95$.
The optimal first-return value is the positive part of that strategy's
value. Guaranteed return alone does not justify maintenance.}
\label{fig:analytical}
\end{figure}

\section{Interpretation, scope, and research questions}
\label{sec:discussion}

\subsection{What Poincar\'e meets Bellman means operationally}

Qualitative organization identifies what can be shared across situations
and which distinctions a reusable interface must preserve. Operational
quotients make that assertion relative to specified future responses and
controlled consequences. Bellman optimization then selects the affordable
path through such organizations, accounting for historical witnesses and
prospective evidence. Recurrence changes continuation value through the
timing and applicability of future demands. The three roles remain
distinct: similarity of dynamics, feasibility of historical repair, and
economic value of retention.

The memory problem includes deciding which distinctions must survive in
some accessible form and which computations can share an active interface.
A split creates new operational addresses. A merge shares addresses or
computation when protected behavior permits sharing. Witness retention
keeps selected future refinements reachable. Forgetting removes repair
options under fixed access. Those decisions generalize the control space
of a conventional fixed-state Bellman problem; the principle of optimality
itself remains unchanged.

One architectural interpretation separates a persistent \emph{scaffold}
of cells, routers, and reusable mechanisms from ongoing \emph{flow}
within those interfaces. Local flow adapts under the deployed organization;
consolidation installs an evidence-supported change to the organization.
The finite catalogs can represent several operating timescales and charge
their storage, computation, and revision costs. Any optimal timescale
separation is conditional on that catalog and cost model. The derivation
does not force one universal scaffold--flow architecture.

An SER allocation can handle specified changing demands without assuming
that one trajectory represents the entire population. The price is an
explicit conditional model, sufficient planning variables, and finite
resource limits. Open-ended adaptation requires choices about tolerated
loss, uncertain future demands, and irrecoverable history. Neither
recurrence nor a hierarchy supplies those choices automatically.

\subsection{How generalization becomes revisable}

Generalization uses a retained rule beyond the examples that produced the
rule. A protected quotient states where such reuse is intended to be valid.
When fresh evidence exposes an exception, local reopening changes the
relevant cell or module while inherited interfaces preserve earlier
responses. The repair succeeds only when the learner can recover the new
address or parameters from its actual memory and admissible access.
Constructibility, preserved competence, and prospective benefit are
separate conditions.

Factor sharing makes combinations economical under structured response
families. Additional independent channels can strengthen evidence under
a persistent regime. Arbitrary response tables and independent historical
demands retain their worst-case information requirements. A rigorous
``blessing of dimensionality'' claim needs a declared factorization or
informative-channel law and a favorable complete resource balance.

\subsection{Alignment as a declared protection problem}

Within the finite model, protected responses can include constraints
representing a supplied alignment specification. A merge must preserve
those responses, and a local split can inherit them at every child address.
An audit can then test a proposed performance improvement while the
migration condition checks implementation. The resulting guarantee is
conditional on the specification, action coverage, observation law, and
prospective drift envelope.

Agreement with a supplied response family does not establish agreement
with unmodeled human values. A task-performance certificate need not
certify safety under a new action-dependent distribution. The finite
framework offers a place to state protection and evidence obligations;
choosing and validating the alignment specification remains a separate
problem.

\subsection{Boundaries and next tests}

The main limitations suggest concrete research tasks. Operational
equivalence needs statistical certification when response laws are unknown,
with action coverage and approximation errors included. The access model
should be learned or robustly bounded without assuming that future evidence
recovers historical episodes. Large catalogs require scalable approximations
to the prescription problem and a measured gap against small exact
benchmarks. Factor and hierarchy discovery need costs for unsuccessful
hypotheses, routing, inherited heads, and audit registers.

A stronger empirical test would expose an agent to recurring contexts,
controlled exceptions, and inaccessible earlier records. Matched complete
budgets should compare distortion-based retention, present-task retention,
and retention chosen for supported repair. Measurements should include
protected-response errors, future target reachability, total stored bits,
computation, audit delay, and net continuation value. Varying the return
law and access channel independently would distinguish reuse frequency
from historical recoverability. Those agent experiments remain future
work; the package currently supplies exact finite checks and analytical
illustrations.

\bibliographystyle{plainnat}
\bibliography{references}
\end{document}